\documentclass[10pt,conference,a4paper]{IEEEtran}
\usepackage[numbers]{natbib}
\usepackage{comment}
\usepackage{graphicx}
\usepackage[cmex10]{amsmath}
\usepackage{amssymb}
\usepackage{amsopn}
\DeclareMathOperator{\med}{med}
\usepackage{array,multirow}
\usepackage{color}
\usepackage{url}

\definecolor{darkgreen}{rgb}{0,0.5,0}
\definecolor{orange}{rgb}{1,0.4,0}

\newcommand{\chk}{\checkmark}

\makeatletter
\def\ps@IEEEtitlepagestyle{%
  \def\@oddfoot{\mycopyrightnotice}%
  \def\@evenfoot{}%
}
\def\mycopyrightnotice{%
{\parbox{18cm}{\vspace{5mm}\footnotesize This manuscript has been accepted for the 23rd International Conference on Pattern Recognition (ICPR 2016)\vspace{2mm}\\  \textcopyright 2016 IEEE. Personal use of this material is permitted. Permission from IEEE must be obtained for all other uses, in any current or future media, including reprinting/republishing this material for advertising or promotional purposes, creating new collective works, for resale or redistribution to servers or lists, or reuse of any copyrighted component of this work in other works.}}
\gdef\mycopyrightnotice{}
}

\begin{document}
\title{Efficient Volumetric Fusion of Airborne and Street-Side Data for Urban Reconstruction}

\author{
\IEEEauthorblockN{Andr\'{a}s B\'{o}dis-Szomor\'{u}}
\IEEEauthorblockA{Computer Vision Lab, ETH Zurich}
\and
\IEEEauthorblockN{Hayko Riemenschneider}
\IEEEauthorblockA{Computer Vision Lab, ETH Zurich}
\and
\IEEEauthorblockN{Luc Van Gool}
\IEEEauthorblockA{Computer Vision Lab, ETH Zurich}
}
\maketitle

\begin{abstract}
Airborne acquisition and on-road mobile mapping provide complementary 3D information of an urban landscape: the former acquires roof structures, ground, and vegetation at a large scale, but lacks the facade and street-side details, while the latter is incomplete for higher floors and often totally misses out on pedestrian-only areas or undriven districts. 
In this work, we introduce an approach that efficiently unifies a detailed street-side Structure-from-Motion (SfM) or Multi-View Stereo (MVS) point cloud and a coarser but more complete point cloud from airborne acquisition in a joint surface mesh. We propose a point cloud blending and a volumetric fusion based on ray casting across a 3D tetrahedralization (3DT), extended with data reduction techniques to handle large datasets. To the best of our knowledge, we are the first to adopt a 3DT approach for airborne/street-side data fusion. Our pipeline exploits typical characteristics of airborne and ground data, and produces a seamless, watertight mesh that is both complete and detailed. Experiments on 3D urban data from multiple sources and different data densities show the effectiveness and benefits of our approach.
\end{abstract}
\IEEEpeerreviewmaketitle

\section{Introduction}

\begin{figure}[!ht]
\centering
\includegraphics[width=40mm, keepaspectratio]{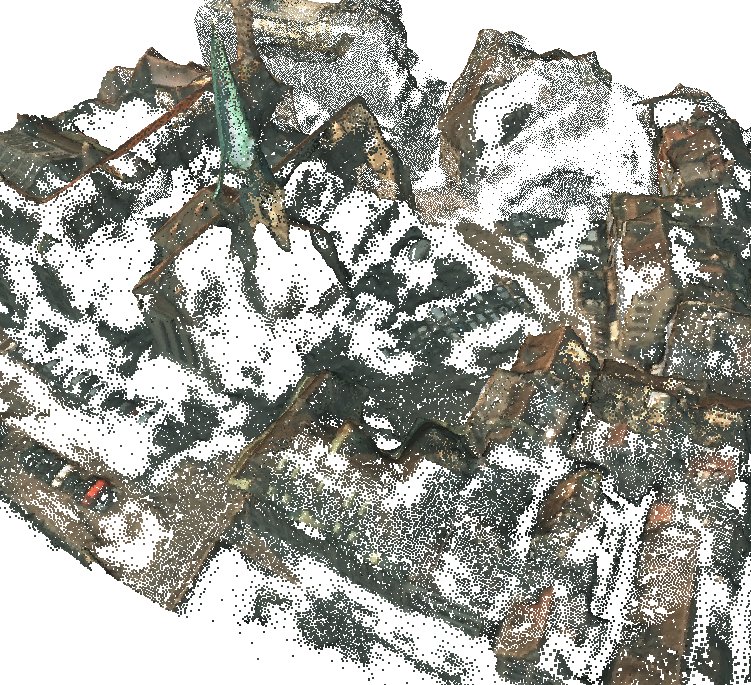}
\includegraphics[width=40mm, keepaspectratio]{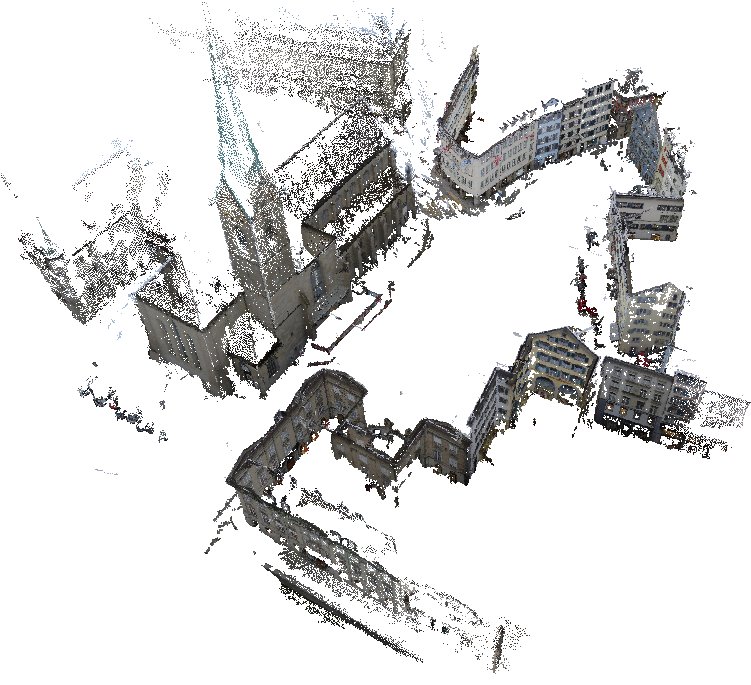}
\includegraphics[width=40mm, keepaspectratio]{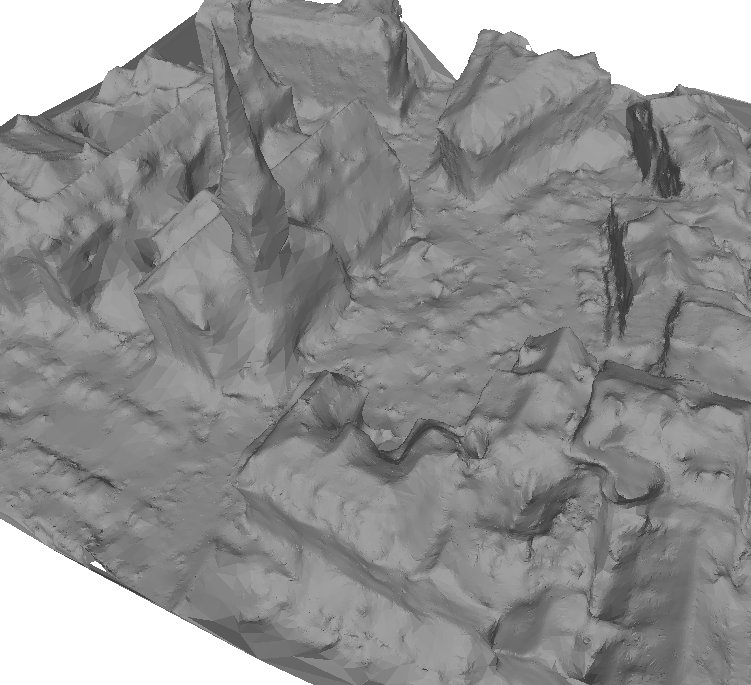}
\includegraphics[width=40mm, keepaspectratio]{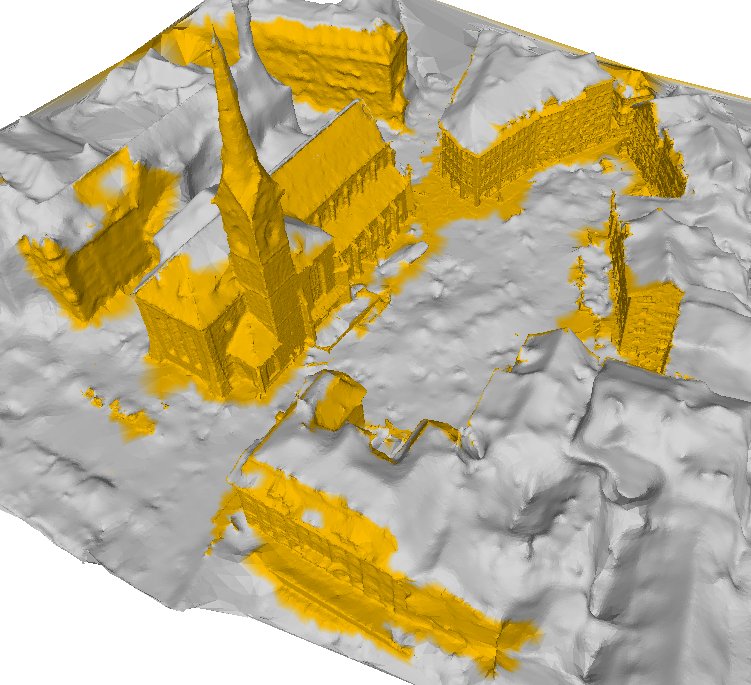}
\includegraphics[width=40mm, keepaspectratio]{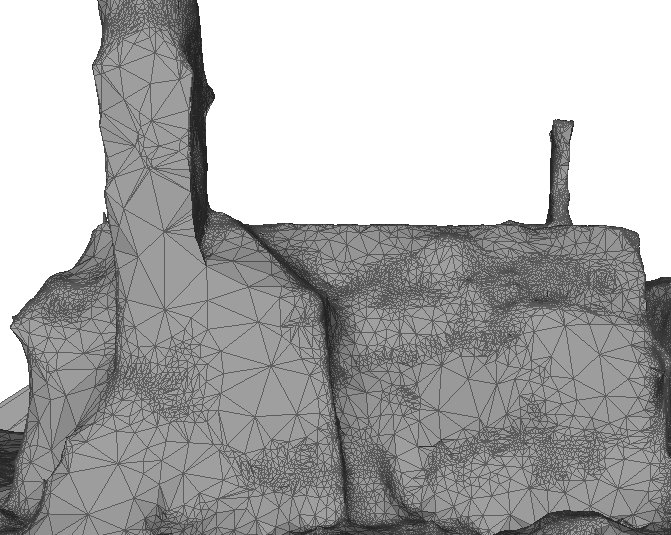}
\includegraphics[width=40mm, keepaspectratio]{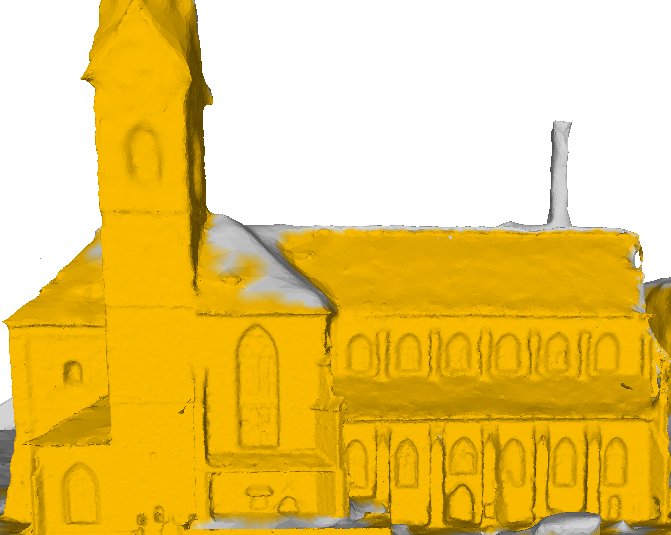}
\vspace{-3mm}
\caption{Our method fuses a complete but inaccurate airborne point cloud (top left) and a detailed but incomplete street-side point cloud (top right) into a joint watertight mesh (right). In comparison, our mesh reconstructed solely from airborne nadir data is shown on the left.}
\label{fig:teaser}
\vspace{-5mm}
\end{figure}

Structure-from-Motion (SfM) techniques have recently been employed at unprecedented scales \cite{HeinlyCVPR2015} to jointly reconstruct outdoor scenes and geo-locate images. However, the resulting point cloud is fragmented and inhomogeneous. 
Multi-View Stereo (MVS) methods are often used to compute a dense surface from known views.
Advances in MVS have enabled the automated production of large-scale urban models from airborne imagery
and it has become a solid alternative to airborne LiDAR.
Unfortunately, such models still often lack facade and street-level details due to occlusions and shadows. 
In turn, MVS has also been applied to street-side mapping \cite{MerrellICCV2007,TolaMVA2011}, as an alternative to LiDAR mobile mapping \cite{FruehIJCV2005}. Automated mobile mapping solutions are prepared to deliver facade details at city-scale, but have no coverage of roofs, higher floors (e.g.~in narrow streets), or traffic-free areas such as courtyards (see Figure~\ref{fig:teaser}). Attempts to model and abstract buildings from such data typically result in facade models much like a floating Potemkin village, i.e. with roofs and ground missing \cite{FruehIJCV2005,RiemenschneiderECCV2014,StrechaCVPR2010}.
We conclude that both airborne and mobile mapping data needs to be exploited in order to produce the next generation of large-scale city models that are to be both complete and detailed. 

In this paper, we propose a solution to fuse an airborne point cloud of large coverage but possibly low detail and a detailed but incomplete street-side point cloud in a volumetric fashion. It is a two-step approach: a mutually exclusive point cloud blending based on graph-cut segmentation, followed by a volumetric optimization over a tetrahedral space-partitioning and based on lines of sight to the points, inspired by~\cite{LabatutCGF2009,LafargeEG2013}. We introduce several simplifications to the volumetric reasoning in order to cope with larger amounts of data, and we show that a drastic reduction of the number of visibility rays and limiting the range of ray-shooting have little effect on reconstruction quality, while the computational workload is reduced significantly. In the same vein, we carry out experiments on different input densities: sparse SfM and dense MVS data. The result is a watertight mesh that is complete at a large scale, but also detailed where street-side data is available. 

Our contributions are the followings. (1) To the best of our knowledge, we are the first to apply volumetric fusion based on a 3D tetrahedralization (3DT) for joint airborne-streetside urban modeling. We propose (2) a mutually exclusive point cloud blending to cope with gross ray conflicts prior to 3DT-fusion (Sect.~\ref{sec:blending}) and (3) techniques to reduce the computational workload for larger scenes (Sect.~\ref{sec:simplify}). Finally, we provide (4) detailed experimentation focusing on the trade-off between computational workload and surface detail (Sect.~\ref{sec:experiments}).

\section{Related Work}\label{sec:related}

Many approaches have been proposed to convert a series of depth maps (range images) or a point cloud -- obtained from SfM, MVS or LiDAR -- into a consistent surface mesh.

Explicit methods directly construct mesh faces over the input points
or depth data or join multiple meshes by zippering them, e.g.~\cite{FruehIJCV2005,FruhCGA2003,MerrellICCV2007}. These are often interpolatory, sensitive to noise and can result in open meshes with holes and non-manifold areas.
Implicit methods extract the surface as a level-set of a volumetric function evaluated over a spatial grid. The seminal work \cite{CurlessSIGGRAPH1996} integrates range images into a voxel grid via a weighted average over Truncated Signed Distance Functions (TSDF).
It has inspired many indoor fusion techniques, e.g.~KinectFusion \cite{IzadiSIGGRAPH2011}. 
Poisson reconstruction \cite{KazhdanSGP2006} is another popular approach due to its robustness to noise.
In general, SDF-based methods tend to oscillate around noisy input. Thus, \cite{HornungSGP2006,LempitskyCVPR2007} drop the sign of the distance function and find the surface as an s/t-cut in a graph over voxels, with regularization for a smooth result. 
Both only show results for small objects. 
Convex variational methods have also been applied for fusing noisy depth maps over a 3D voxel grid \cite{ZachICCV2007} or a 2.5D height map \cite{Pock2011}. 

Volumetric reasoning over a voxel grid is prohibitively expensive for large outdoor scenes, motivating adaptive space partitioning methods. Octrees are used in Dual Contouring \cite{JuSIGGRAPH2002} and Poisson reconstruction \cite{KazhdanSGP2006} but 3D Delaunay-tetrahedralization (3DT) \cite{LabatutICCV2007,VuPAMI2012}, and a cell complex built from a plane arrangement \cite{ChauveCVPR2010} have also been applied to such scenes. Labatut et al.~\cite{LabatutICCV2007} build a 3DT on top of MVS points and do inside/outside classication of tetrahedra while enforcing line-of-sight and photoconsistency constraints. 
Later, the expensive photoconsistency term was dropped, and a remarkable robustness to outliers and to subsampling was shown \cite{LabatutCGF2009}. Their street-side results appear superior in quality compared to TV-L$^1$ fusion \cite{ZachICCV2007}. As a drawback, the output mesh is interpolatory with the noisy input points as vertices. 
The 3DT approach has been extended by a photoconsistency-driven mesh optimization \cite{VuPAMI2012}, and it has been adapted to point clouds with no visibility data by casting ray ``tubes" \cite{LafargeEG2013}. Another extension addressing weakly-supported surfaces is implemented in CMP-MVS \cite{JancosekCVPR2011}, which outputs a good-quality mesh in a matter of hours from hundreds of images of a street scene on a 3~GHz computer with GPU.  
It shall be noted that none of these techniques are specific to fusing aerial and street-side data. 

Recently, we proposed an efficient view-driven meshing approach for street-side images and for large-scale height maps \cite{BodisCVIU2016}.
The benefits of combining different data sources for urban modeling have also been recognized by others \cite{StrechaCVPR2010,CabezasCVPR2014}. 
However, surface \emph{fusion} of street-side and aerial data has received significantly less attention to date. Many works focus on geo-localization of either images or street-side reconstructions \cite{StrechaCVPR2010,KaminskyWS2009,Shan3DV2014}, a prerequisite for fusion. Their input is community photo collections, which are spatially fragmented, i.e.~only represent popular landmarks. 
In turn, systematic industrial airborne/street-side mobile mapping deliver denser coverage and geo-registered data by means of GPS/IMU sensors and/or ground control points.
Therefore, our approach assumes that the input is geo-registered (aligned) and focuses on the surface reconstruction step that tolerates minor misalignments.

Last but not least, there exist only a few works that combine street-side and aerial data for joint mesh reconstruction \cite{FruhCGA2003,Fiocco3DIM2005,Shan3DV2013}. Fr\"uh and Zakhor \cite{FruhCGA2003} construct meshes over street-side LiDAR range maps and over a large-scale Digital Surface Model (DSM). 
Unlike in our approach, they reconstruct a facade and an airborne mesh separately without topological fusion.
Shan et al. \cite{Shan3DV2013} solves the problem by directly applying Poisson surface reconstruction \cite{KazhdanSGP2006} over the joint dense point cloud computed by patch-based MVS \cite{FurukawaPAMI2010} without a cross-consistency check between airborne and street-side data. Fiocco et al.~\cite{Fiocco3DIM2005} integrate over 200M points from a tripod-mounted ground LiDAR and an aerial DSM, by using a distance field over an octree and an out-of-core dual contouring approach. Despite the quite complex algorithm, results are noisy and contain many large holes. In contrast, our approach performs cross-consistency filtering and produces a good quality watertight surface.

\section{The Proposed Method}\label{sec:method}
In what follows, we first detail the volumetric fusion method, then propose an \emph{a priori} point cloud blending, and finally discuss how we reduce the computational effort.

\subsection{Volumetric fusion}\label{sec:fusion}

Inspired by the elegance of tetrahedral space partitioning methods \cite{LabatutCGF2009,LafargeEG2013}, we use a 3D Delaunay-tetrahedralization (3DT) over the joint point cloud from airborne and street-side acquisition (with the modifications discussed later). The underlying data structure is simpler than an octree or cell complex, yet it is adaptive to data density, a prerequisite for scalability. Denote the vertex set of the 3DT as $\mathcal{V}=\{v_i\}$, the set of tetrahedra by $\mathcal{T}=\{t_i\}$, and the triangular facet between any adjacent tetrahedra $t_i$ and $t_j$ as $f_{ij}$. Our goal is to assign a binary label $l_i\in\{in,out\}$ to every tetrahedron $t_i\in\mathcal{T}$, while minimizing the energy function
\begin{align}
E(\mathcal{L})=
\sum_{i:t_i\in\mathcal{T}}E_i(l_i)+
\sum_i\sum_{j:j<i}E_{ij}\cdot\mathbb{I}[l_i\neq l_j],
\label{eq:energy}
\end{align}
where the unary term $E_i(l)$ encodes the preference for tetrahedron $t_i$ to obtain label $l$, the pairwise regularization term $E_{ij}$ is the preference for two adjacent tetrahedra $t_i$ and $t_j$ to obtain the same label, $\mathbb{I}[l_i\neq l_j]$ is the indicator (1 if $l_i\neq l_j$, 0 otherwise), and $\mathcal{L}=(l_1,l_2,\dots,l_n)$ is a complete labelling.

\begin{figure}[!ht]
\centering
\includegraphics[height=25mm, keepaspectratio]{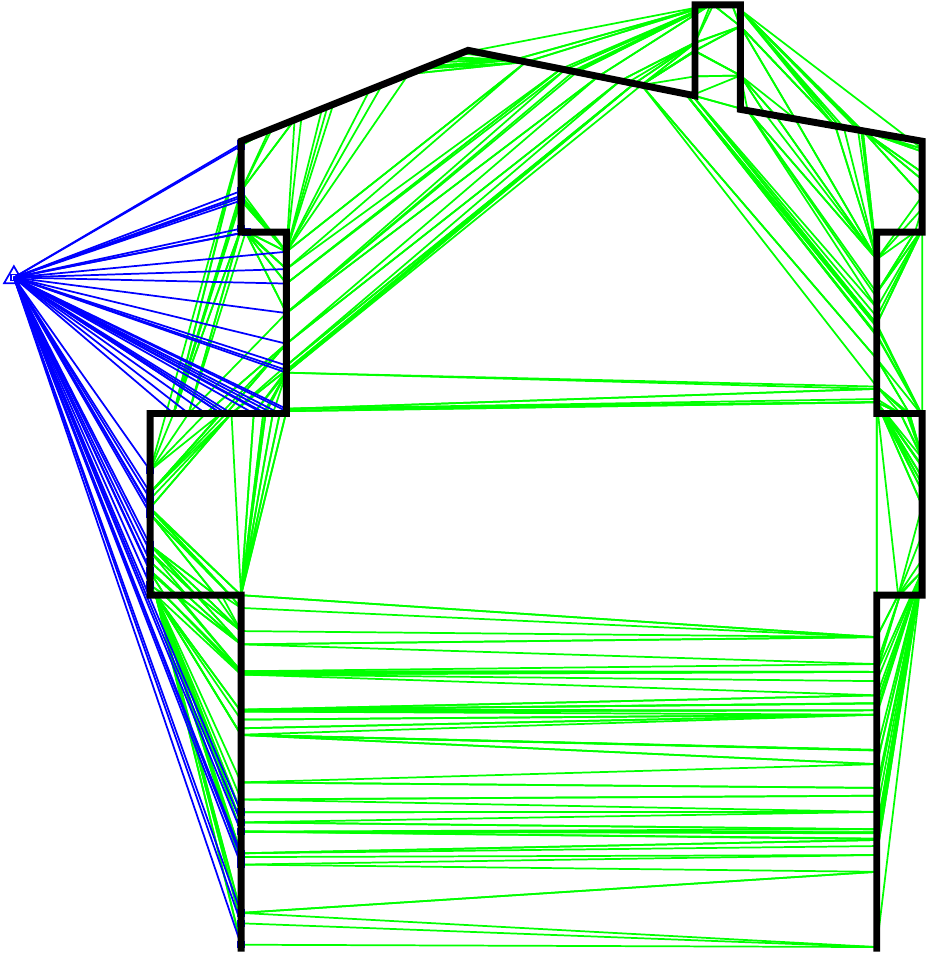}
\includegraphics[height=25mm, keepaspectratio]{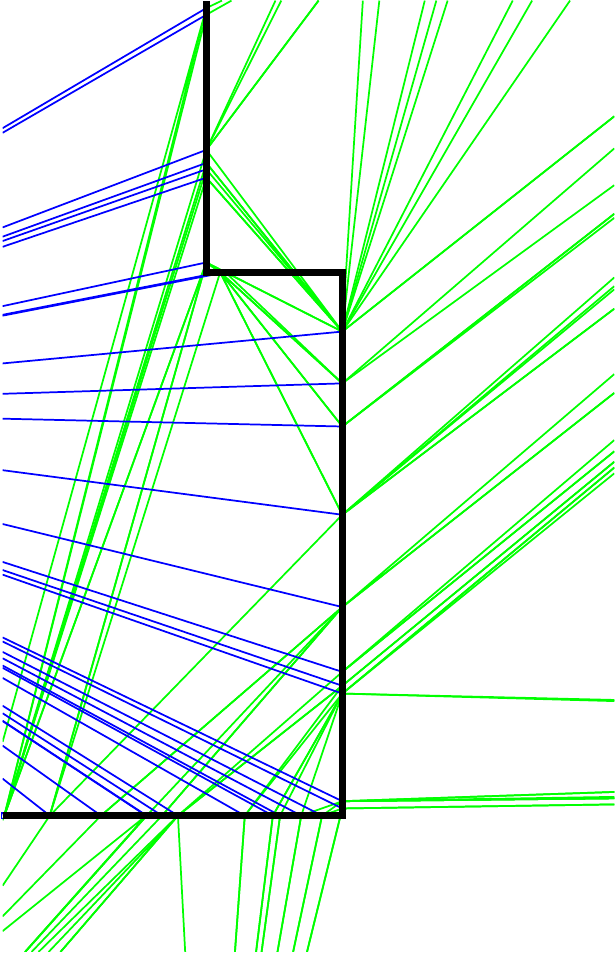}
\includegraphics[height=25mm, keepaspectratio]{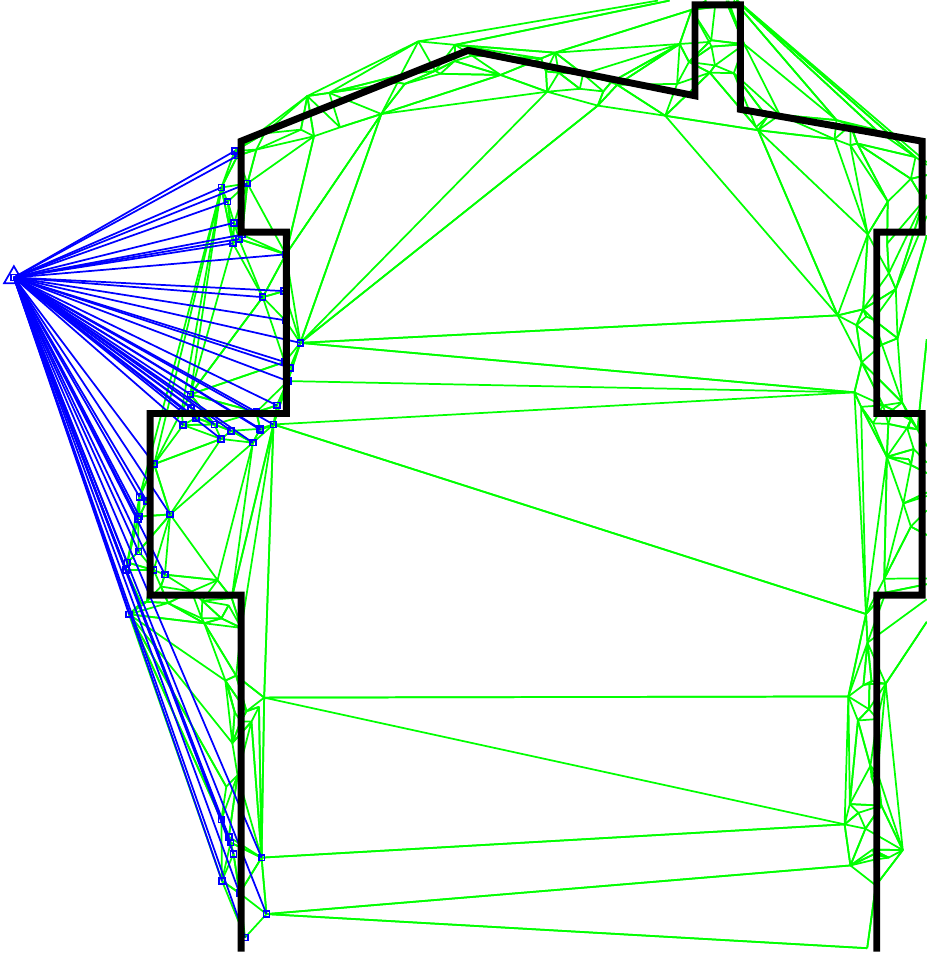}
\includegraphics[height=25mm, keepaspectratio]{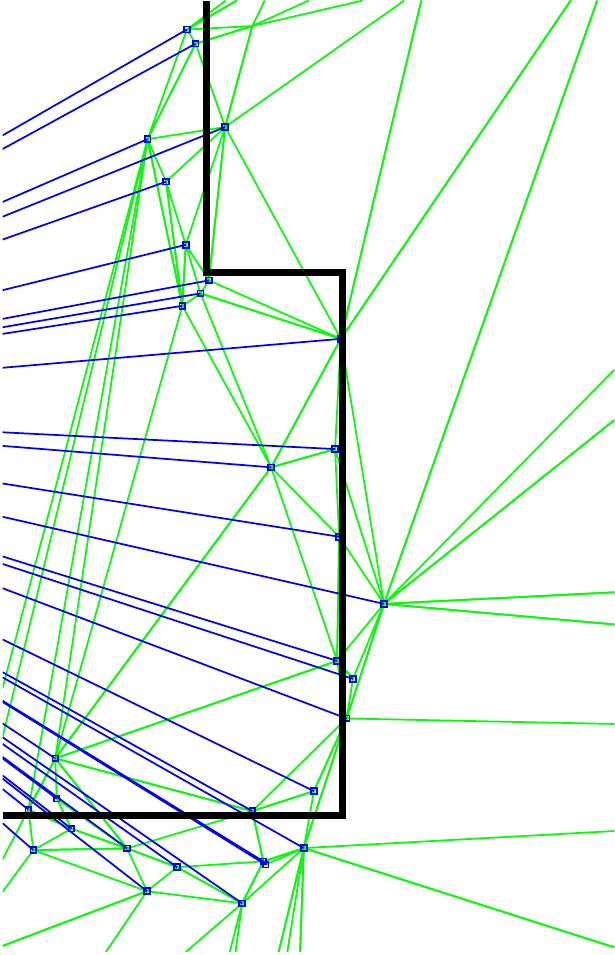}
\vspace{-2mm}
\caption{2D illustration of the effect of noise on tetrahedralization (green) and on the visibility rays (blue) to a single sensor. Point samples on the true surface (black) are Delaunay-triangulated without (left) and with noise (right). Close-ups are also shown. Noise increased the number of triangles by 8\%.}
\label{fig:simu2d}
\end{figure}

As input, we construct lines of sight (rays) from the visibility information of the SfM or MVS (or laser scanning) procedure. Every ray $r=(v,s)$ goes from a vertex $v\in\mathcal{V}$ to a sensor location $s$. The logic behind inside/outside reasoning is that tetrahedra crossed by rays should be labelled as $out$, while tetrahedra behind each ray's vertex of origin should be labelled as $in$ up to a certain distance $\delta_{max}^{in}$. This distance should be conservative to avoid mistakes around corners and behind narrow objects. In practice, rays either penetrate or do not reach the true surface due to noise as shown in Fig.~\ref{fig:simu2d}.

\begin{figure}[!ht]
\centering
\includegraphics[width=85mm, keepaspectratio]{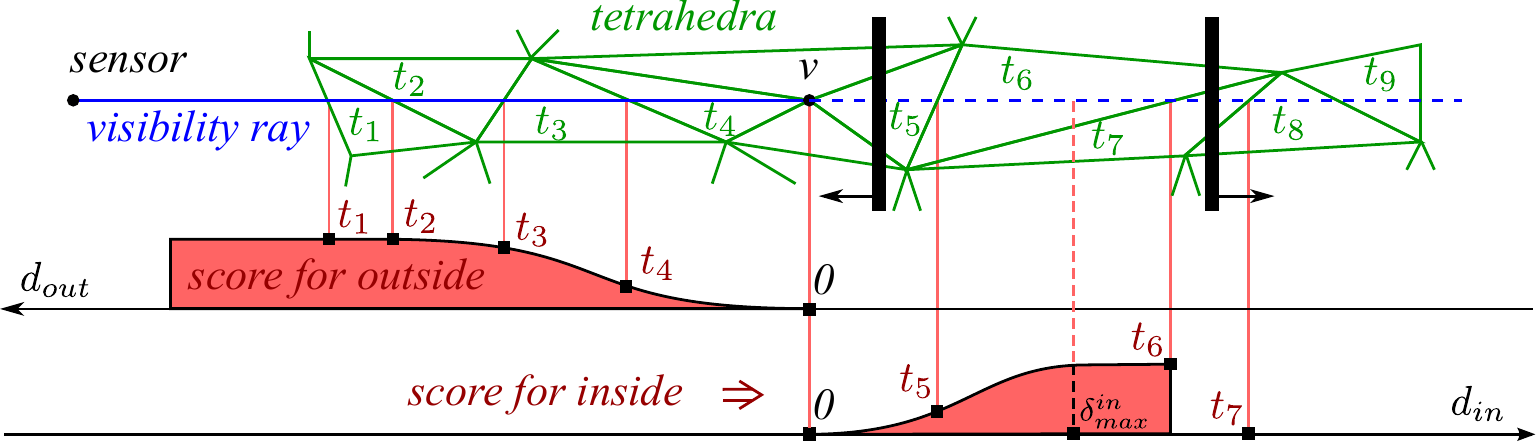}
\vspace{-2mm}
\caption{Penalties arising from a single ray (solid blue) originated from vertex $v$. The true surface (black lines) is oriented towards outside (arrows). Vertices are off-surface due to noise. Crossed tetrahedra $\{t_1,t_2,t_3,t_4\}$ have a preference for $l=out$, while $\{t_5,t_6\}$ have preference for $l=in$.}
\label{fig:ray}
\end{figure}
Each ray contributes to the unary terms $E_i(l)$ in Eq.~\eqref{eq:energy}. For $l=out$, we follow the path of each ray $r$ by walking in the 3DT (Fig.~\ref{fig:ray}). The walking procedure starts from the source vertex $v$ of the ray, returns each traversed tetrahedron $t$, and the walking distance to the point where the ray exits $t$. The walk ends with the tetrahedron containing the sensor location $s$, or where the ray exits the 3DT, whichever occurs first. For $l=in$, the inverted ray is traced (dotted blue line in Fig.~\ref{fig:ray}) until the tetrahedron at distance $\delta_{max}^{in}$ from $v$. 

In order to take noise (Fig.~\ref{fig:simu2d}) into account, we employ a soft-voting scheme, similar to \cite{LabatutCGF2009}. 
Denote the exit distance of ray $r$ from tetrahedron $t$ by $d_l(r,t)$, where $l\in\{in,out\}$ distinguishes the two walking directions. $d_l(r,t)=0$ for tetrahedra not affected by ray $r$.
For a single ray $r$, the score of any tetrahedron $t$ to be labelled as $l\in\{in,out\}$ is
\begin{align}
	S_l(r,t)=1-e^{-d_l^2(r,t)/(2\sigma_l^2)},\label{eq:margin}
\end{align}
where $\sigma_{in}$ and $\sigma_{out}$ are noise tolerance parameters. The truncation distance is fixed to $\delta_{max}^{in}=3\sigma_{in}$, i.e. the last tetrahedron traversed backwards obtains the full score 1 for $l=in$ (see Fig.~\ref{fig:ray}).
The penalty $E_i(l)$ of tetrahedron $t_i$ for label $l$ is then defined as a monotonic function of the sum of the preferences of the opposite label $\bar{l}$ over all rays:
\begin{align}
	E_i(l)=1-e^{-U_i(l)/\gamma_l}\hspace{2mm}\text{with}\hspace{2mm}
	U_i(l) = \sum_r S_{\bar{l}}(r,t_i),
\end{align}
where $\gamma_{in}$ and $\gamma_{out}$ control how many $in$ and $out$ votes collected from all rays per tetrahedron count as ``many enough'', and can be used to trade-off robustness and accuracy. 

The terms $E_{ij}$ in Eq.~\eqref{eq:energy} enforce spatial regularization for a smooth solution, and propagate labels to tetrahedra that have no unary preference. 
$E_{ij}$ is a pairwise term that is a function of properties of two adjacent tetrahedra $t_i$ and $t_j$. In this paper, we show experiments using a simple minimal area force
\begin{align}
	E_{ij}=\lambda A_{ij},
\end{align}
where $\lambda$ is a global regularization factor, and $A_{ij}$ is the area of triangle $f_{ij}$ between tetrahedra $t_i$ and $t_j$. Other forces are also possible: an initial experimentation with forces against elongated triangles, long-edge triangles, or preference for faces $f_{ij}$ between any two elongated tetrahedra \cite{LabatutCGF2009} (see Fig.~\ref{fig:simu2d} for an intuition) shows that the area penalty works best with our datasets. Further comparison is out of the scope of this paper.
Note that the margins introduced by the soft-voting scheme (Fig.~\ref{fig:ray}) for the unary term let the optimization select from a larger pool of faces.

The energy function \eqref{eq:energy} can be globally minimized using graph-cuts \cite{Boykov2001}. The sought optimal surface mesh can be extracted as the set of all faces between inside and outside tetrahedra. Although this mesh is watertight, it is interpolatory, since the 3DT is built on top of noisy points. This results in a jagged surface inherent in most 3DT-based methods. To obtain a smoother surface, we apply basic Laplacian smoothing by replacing every vertex by the mean of its neighbors. Finally, only the largest component is kept in order to eliminate (typically a few) minor components not attached to the ground. 

\subsection{Point cloud blending}\label{sec:blending}
\begin{figure*}[!ht]
\centering
\includegraphics[width=28mm, keepaspectratio]{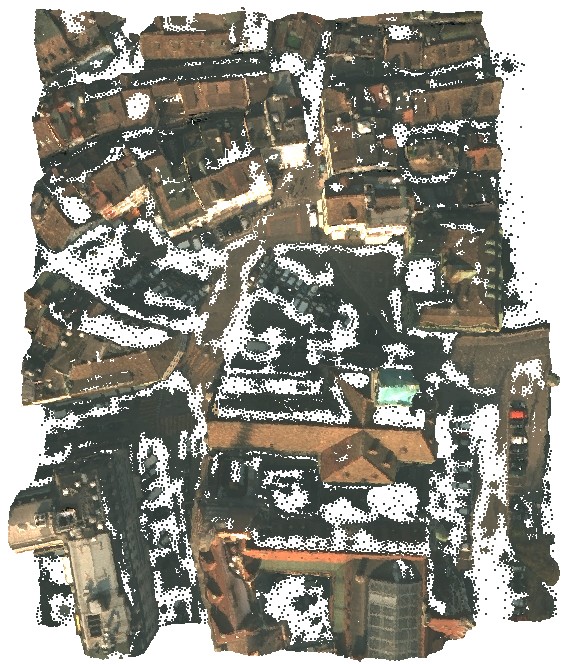}
\includegraphics[width=28mm, keepaspectratio]{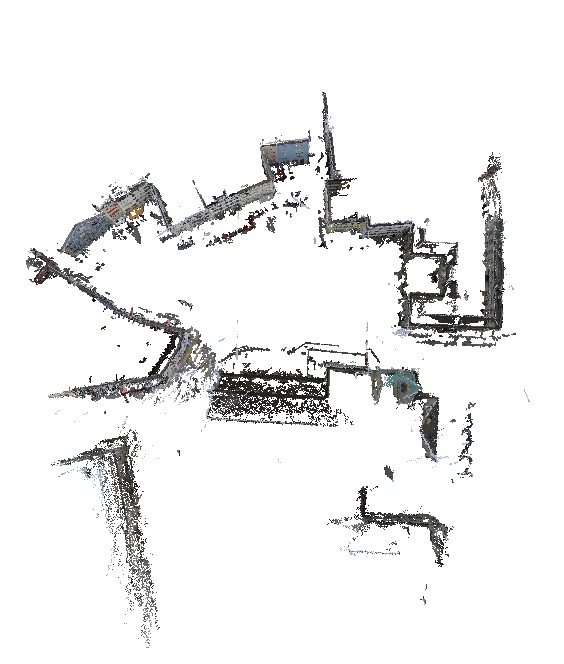}
\includegraphics[width=28mm, keepaspectratio]{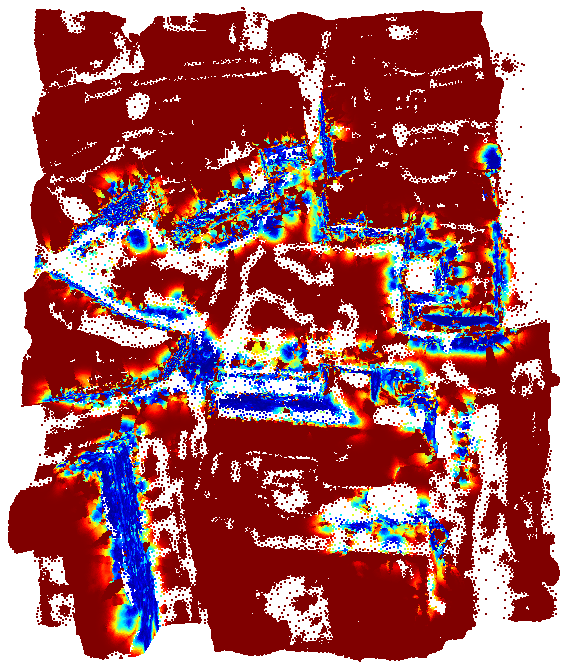}
\includegraphics[width=28mm, keepaspectratio]{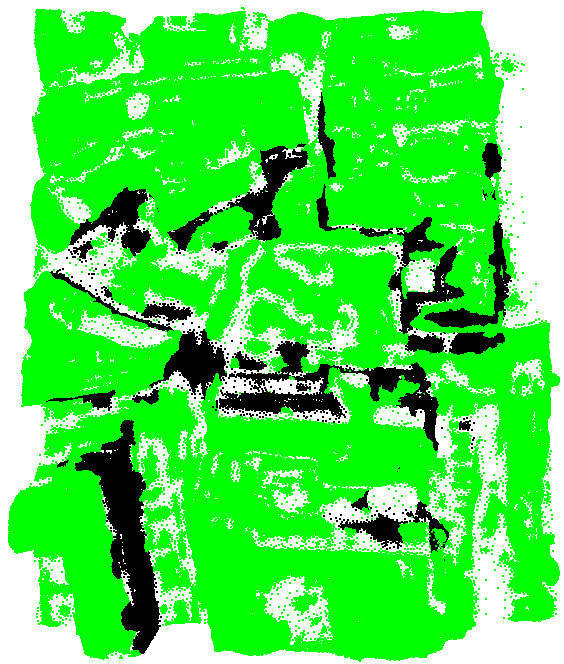}
\includegraphics[width=28mm, keepaspectratio]{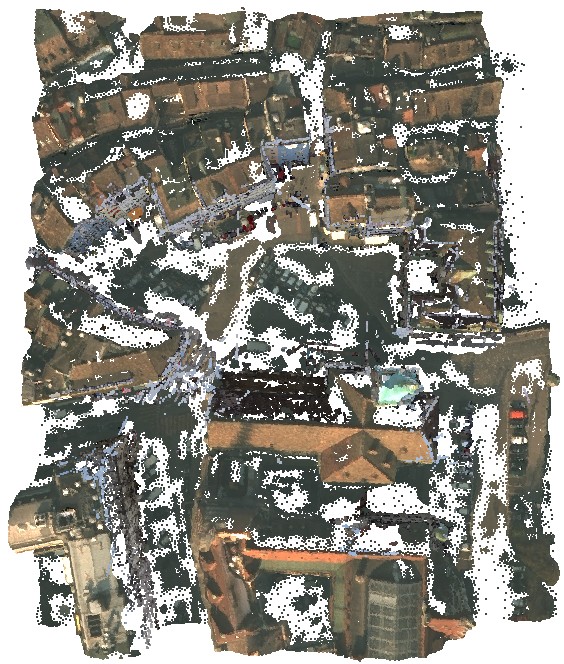}
\includegraphics[width=28mm, keepaspectratio]{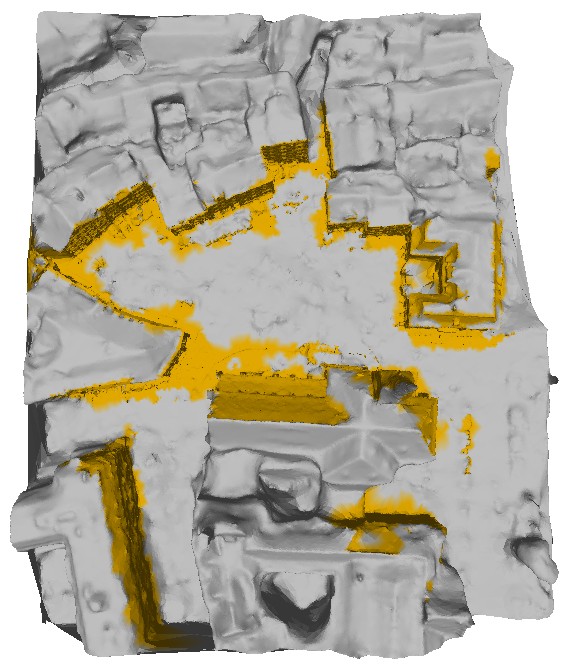}\\
\includegraphics[width=28mm, keepaspectratio]{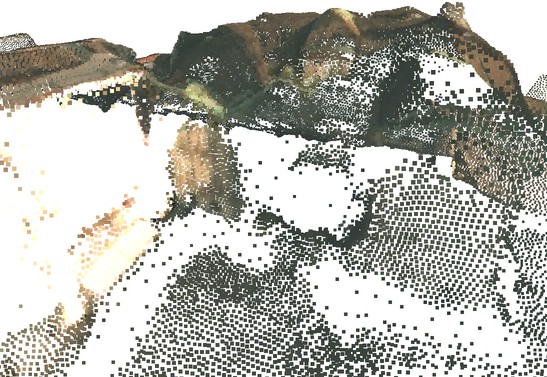}
\includegraphics[width=28mm, keepaspectratio]{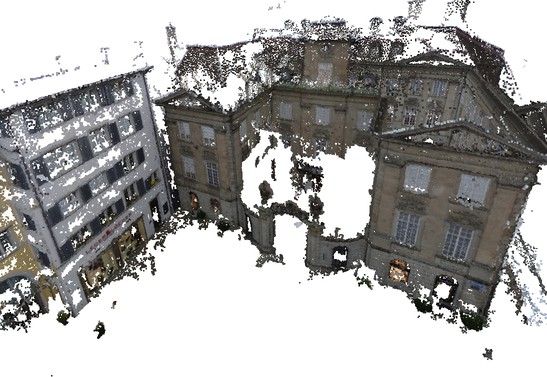}
\includegraphics[width=28mm, keepaspectratio]{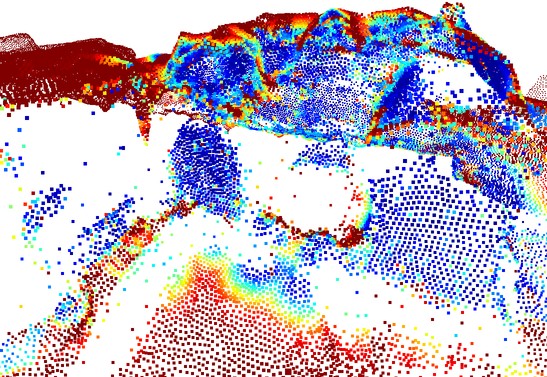}
\includegraphics[width=28mm, keepaspectratio]{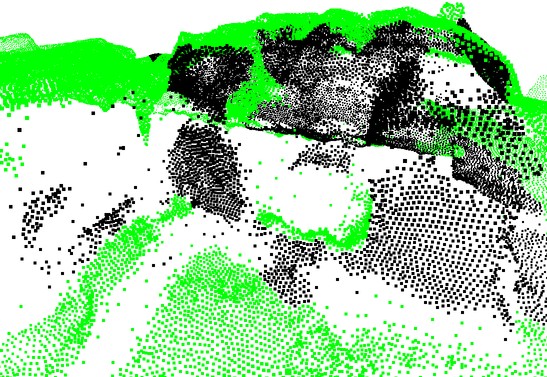}
\includegraphics[width=28mm, keepaspectratio]{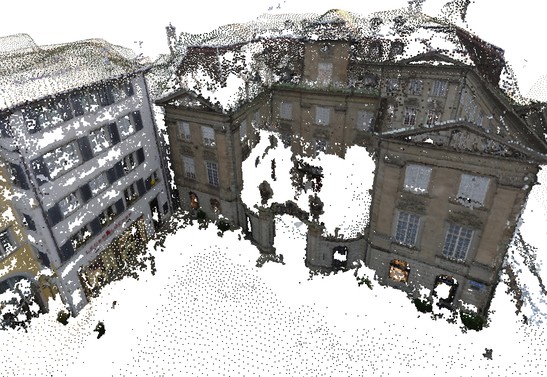}
\includegraphics[width=28mm, keepaspectratio]{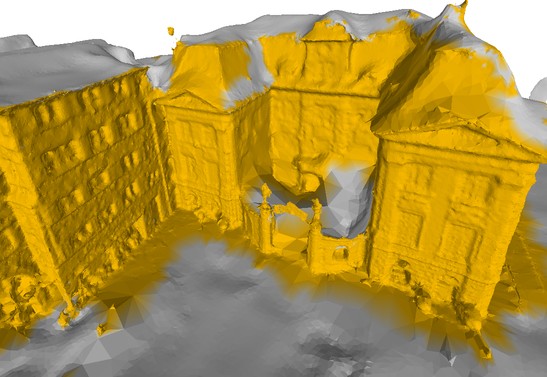}\\
\vspace{-2mm}
\caption{Our pipeline applied to M\"unsterhof (PMVS). Overview (top) and a close-up (bottom). From left to right: input aerial point cloud, input street side point cloud, blending energies, airborne point cloud segmentation, blended aerial and street-side point clouds, output fused mesh with street-side parts in orange.}
\label{fig:overview}
\vspace{-5mm}
\end{figure*}
Unfortunately, applying the proposed volumetric fusion procedure \emph{directly} to the joint airborne/street-side set of points does not give satisfactory results in practice. The main reason is that airborne reconstructions are usually substantially lower quality and oversmoothed at facades, which results in duplicate surfaces and ray conflicts across the two data types. Rays from airborne data often intersect the true surface or show inside evidence far outside of the true surface. Setting our fusion parameters $\sigma_{in},\sigma_{out}$ loose would result in loss of detail. 
To get the best out of both data types, we rather propose a mutually exclusive point cloud blending prior to volumetric fusion.
Besides the considerations that \emph{(i)} the airborne data has broad coverage, \emph{(ii)} street-side input is incomplete or fragmented and \emph{(iii)} airborne data is of substantially lower quality where street-side data is available, it is reasonable to eliminate airborne points where street side data is present. The removal of points comes with the additional computational benefit of lowering the number of tetrahedra in the partitioning. 

We formulate this as a segmentation over the airborne point cloud $\mathcal{P}$, which assigns a binary label $l\in\{0,1\}$ to each point $p_i\in\mathcal{P}$, points to remove being marked by 0. 
Given the street-side point cloud $\mathcal{Q}$, an airborne point $p_i$ has a (supposedly better quality) street-side substitute $q_i\in\mathcal{Q}$ if $q_i$ is the nearest neighbor of $p_i$ in $\mathcal{Q}$ and if both the Euclidean distance $d_i$ between them and the angle $\theta_i$ between their normals $\mathbf{n}(p_i)$ and $\mathbf{n}(q_i)$ are small. The likelihood for an airborne point to have a substitute can be formulated as
\begin{align}
\phi_i=\phi(d_i,\theta_i)=e^{-d_i^2/(2\sigma_b^2)}\cdot \max\{0,\cos\theta_i\},
\end{align}
where $\cos\theta_i=\mathbf{n}(p_i)^T\mathbf{n}(q_i)$, and $\phi$ ranges from 0 (no substitute) to 1 (perfect substitute). $\sigma_b$ is a blending parameter to control our notion of vicinity, which should incorporate deviations of $\mathcal{P}$ from $\mathcal{Q}$ due to registration and reconstruction errors. 
Normals are computed in $\mathcal{P}$ and $\mathcal{Q}$ separately by least-squares plane fitting to the $k$-NN neighborhood of each point ($k=10$), and by flipping normals according to visibility. For a smooth segmentation, we define the influence between adjacent nodes of the $k$-NN graph over $\mathcal{P}$ as
\begin{align}
\psi(p_i,p_j) = \exp(-d_{ij}/\med d_{ij}),
\end{align}
where $d_{ij}$ is the distance between any two adjacent points $p_i$ and $p_j$, and $\med d_{ij}$ is the median of all $k$-NN distances in $\mathcal{P}$.
We seek the binary labelling over $\mathcal{P}$ that minimizes
\begin{align}
	E^b(\mathcal{L})=\sum_{i:p_i\in\mathcal{P}}E^b_i(l_i) + \lambda_b\sum_{ij}\psi(p_i,p_j)\cdot\mathbb{I}[l_i\neq l_j],
	\label{eq:pcl-energy}
\end{align}
where $l_i\in\{0,1\}$ is the label of point $p_i$, $\mathbb{I}$ is the indicator function, and $\lambda_b$ is a regularization parameter. The unary penalties $E^b_i(l)$ for point $p_i$ to obtain label $l$ are defined as
\begin{align}
	E^b_i(0)=1-\phi_i\hspace{2mm}\text{and}\hspace{2mm}E^b_i(1)=\phi_i.
\end{align}

We minimize Eq.~\eqref{eq:pcl-energy} by graph-cuts \cite{Boykov2001} and eliminate the airborne points labelled 0 prior to volumetric fusion. Although point quality measures could be incorporated into this scheme, we have found that the distance and normal cues prove sufficient to obtain a good mixture of the two point clouds.

\subsection{Data reduction}\label{sec:simplify}

Our aim is to apply the pipeline to large scenes. To reduce the computational complexity,  we propose different simplifications, whose effect will be exprimented in Section~\ref{sec:experiments}.

\subsubsection{Input point decimation}
We cluster the blended point cloud using a voxel grid, and replace the points in each voxel by their centroid. The visibility information is also merged accordingly. To keep the method scalable, only occupied voxels (a small fraction) are stored, using an associative array indexed with the global voxel index.

\subsubsection{Reducing the number of rays}
Lines-of-sight are distributed densely in free-space areas, leaving space for a sparsification. We reduce the number of rays per point to one. 
The ray that is closest to the normal direction of each point is kept, to minimize surface penetration due to noise in the points.

\subsubsection{Ray truncation}
When collecting inside votes in the fusion algorithm (Sect.~\ref{sec:fusion}), the penetration depth has intentionally been limited to $\delta_{max}^{in}=3\sigma_{in}$ to preserve narrow structures. We introduce a similar limit $\delta_{max}^{out}=3\sigma_{out}$ when tracing the rays from each vertex $v$ towards the sensor (solid blue line in Fig.~\ref{fig:ray}) for a shorter walk in the 3DT. 
 
\section{Experiments}\label{sec:experiments}
Our mixed C++/Matlab single-core implementation relies on Matlab/CGAL \cite{CGAL} for tetrahedralization, and on the GCOptimization library \cite{Boykov2001} for graph-cuts. We implemented 3DT adjacency computations and ray shooting in C++.  

\begin{figure*}[!ht]
\centering
\vspace{-3mm}
\includegraphics[height=30mm, keepaspectratio]{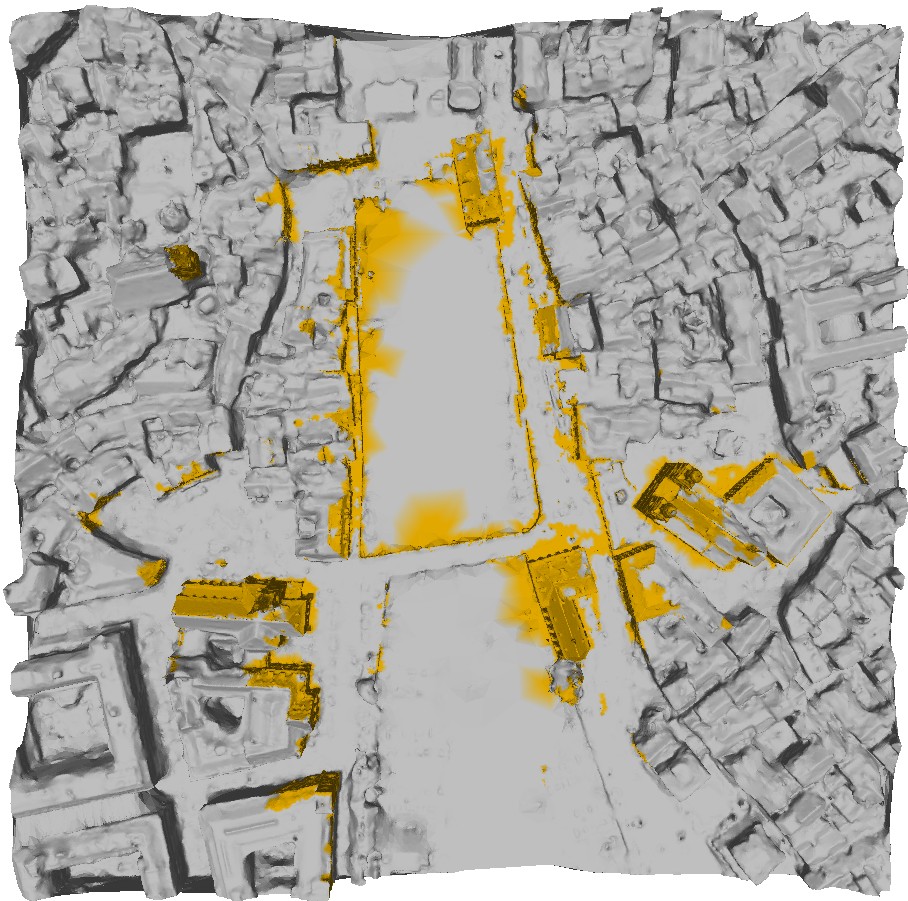}
\includegraphics[height=30mm, keepaspectratio]{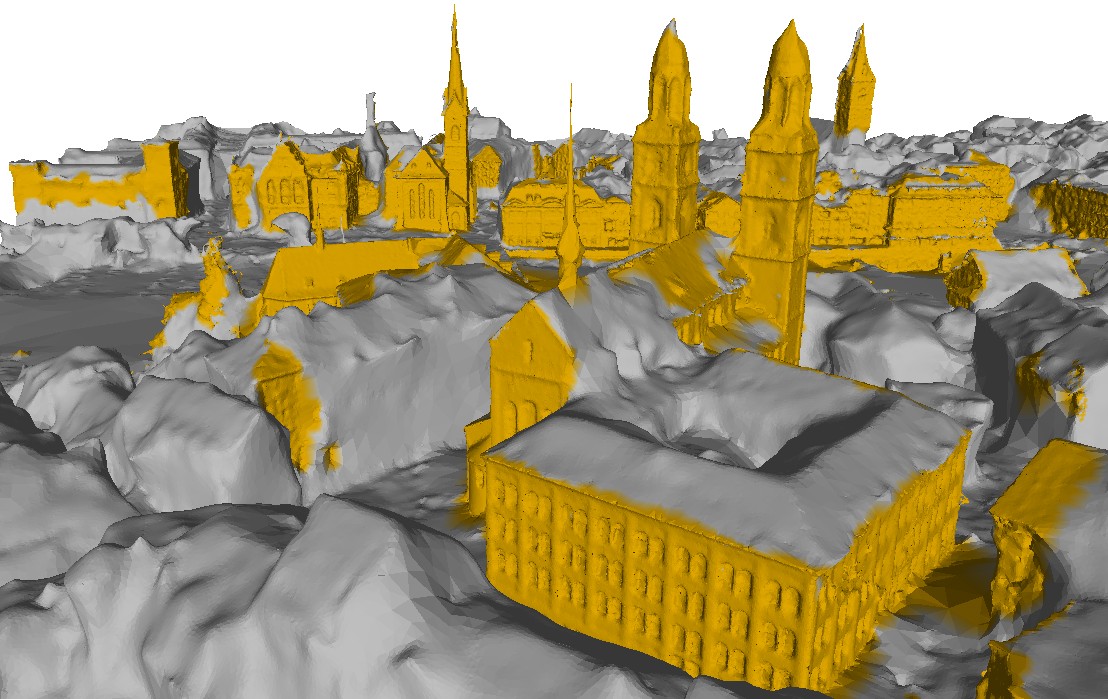}
\includegraphics[height=30mm, keepaspectratio]{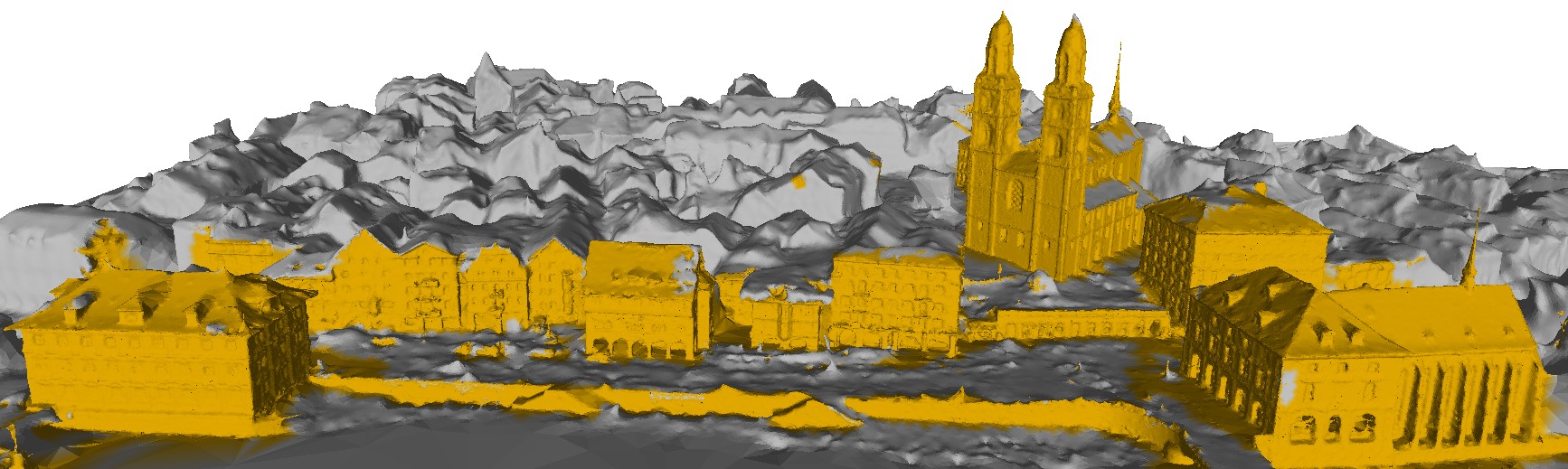}
\vspace{-2mm}
\caption{Output of our fusion for the Limmatquai dataset using PMVS street-side input (see PMVS (*) in Table~\ref{tab:results}). Vertices originated from street-side data are colored orange. The left image shows the full 400$\times$400~m$^2$ result. Around 20\% of surface area is covered by street-side data. }
\label{fig:limmatquai}
\end{figure*}

We are not aware of any publicly available dataset with both airborne and street-side data for the same geographic location. Thus, we show experiments on our datasets, M\"unsterhof (140$\times$160~m$^2$, Fig.~\ref{fig:overview}) and Limmatquai (400$\times$400~m$^2$, Fig.~\ref{fig:limmatquai}) captured in Z\"urich, Switzerland. 
23 nadir images of 15~cm ground resolution, taken at a height of 3~km, were fed to a state-of-the-art MVS pipeline\footnote{RealityCapture, \url{https://www.capturingreality.com/}} to obtain an airborne point cloud. 
629 (M\"unsterhof) and 847 (Limmatquai) street-side images were taken manually in the same areas, and we ran VisualSfM \cite{VisualSfM} to produce an SfM point cloud, as well as PMVS \cite{FurukawaPAMI2010} for a denser cloud. The aerial data was geo-registered via ground control points, and the street-side models were registered manually to the aerial data (assuming that street-side geo-registration can be solved automatically in industrial mobile mapping). To quantify input misalignment (incl. noise), we measure distances between mutual nearest neighbors across the SfM/aerial clouds along the normal directions pointwise, and report the median, 90- and 99-percentiles. These are 0.2, 0.5, 1.0~m for M\"unsterhof and 0.2, 0.6, 1.2~m for Limmatquai.

\begin{figure}[!ht]
\centering
\includegraphics[width=28mm, keepaspectratio]{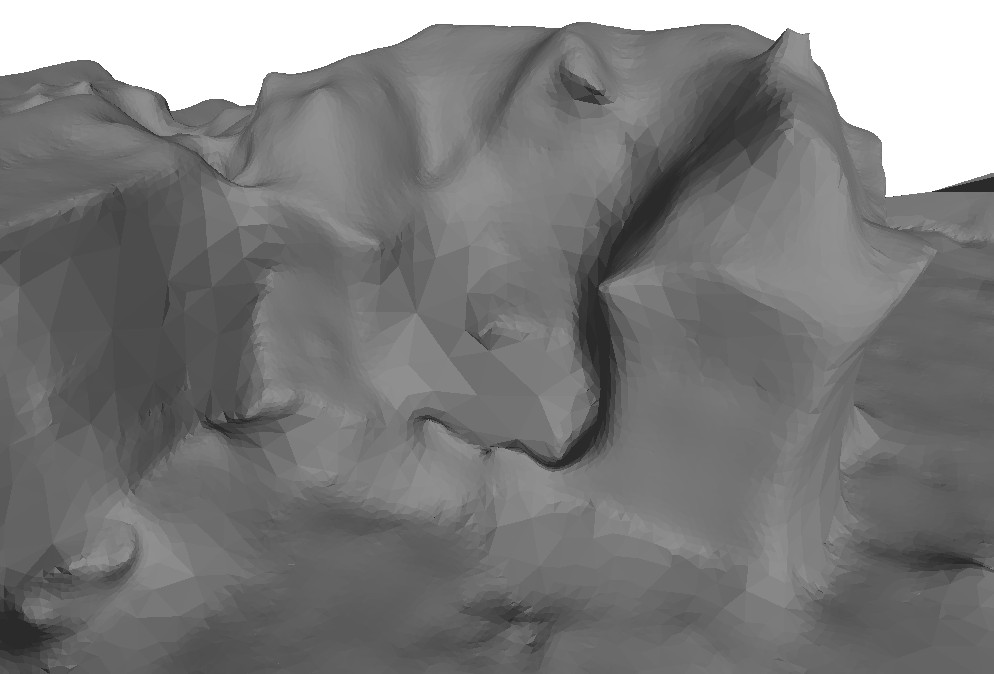}
\includegraphics[width=28mm, keepaspectratio]{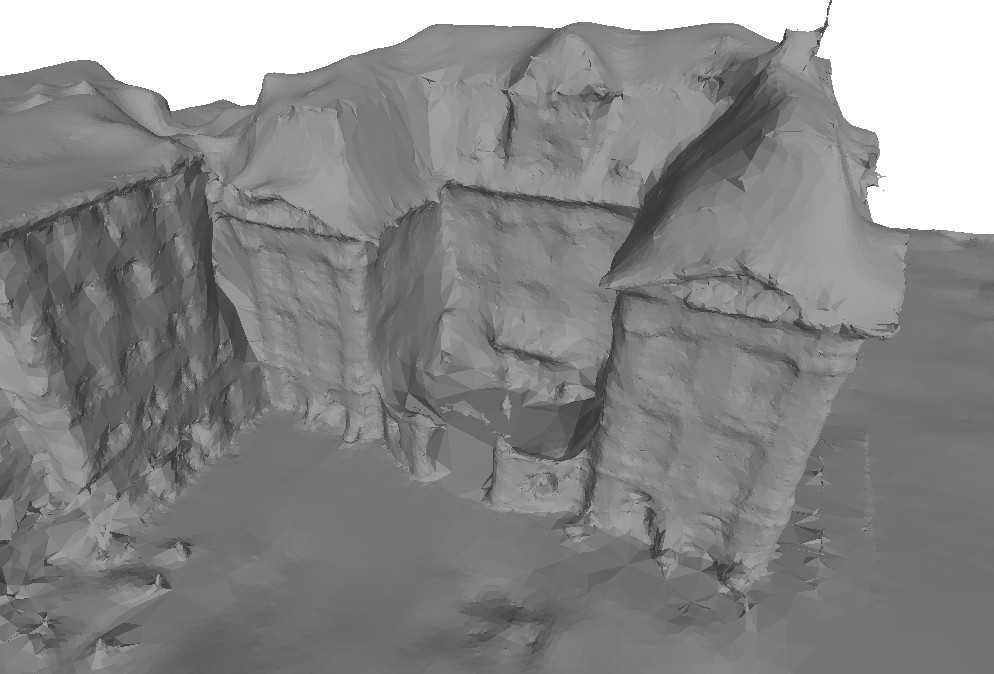}
\includegraphics[width=28mm, keepaspectratio]{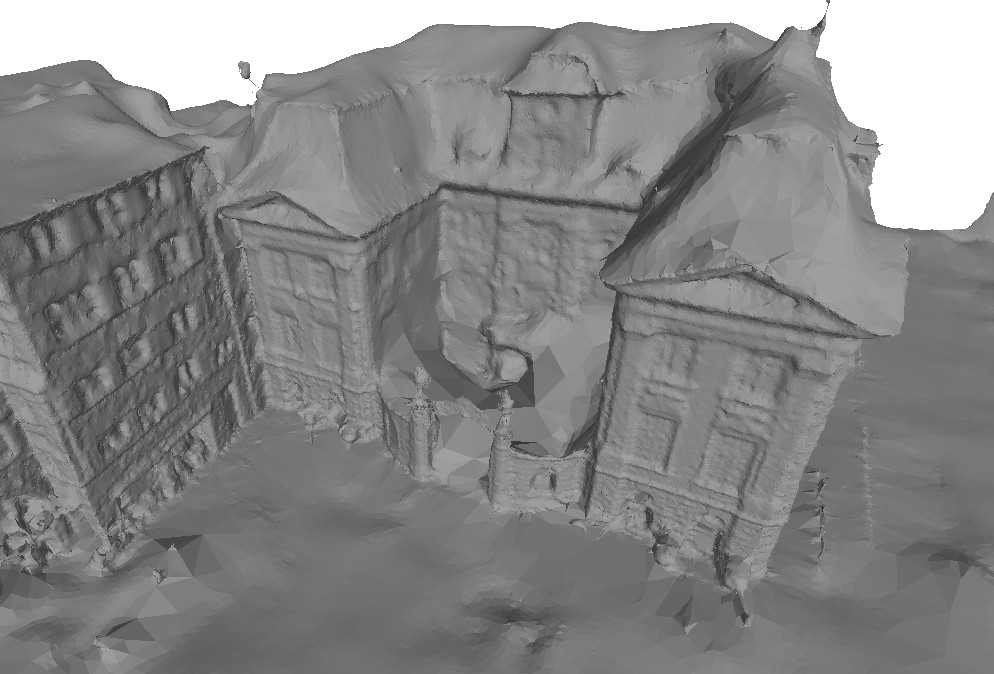}

\includegraphics[width=28mm, keepaspectratio]{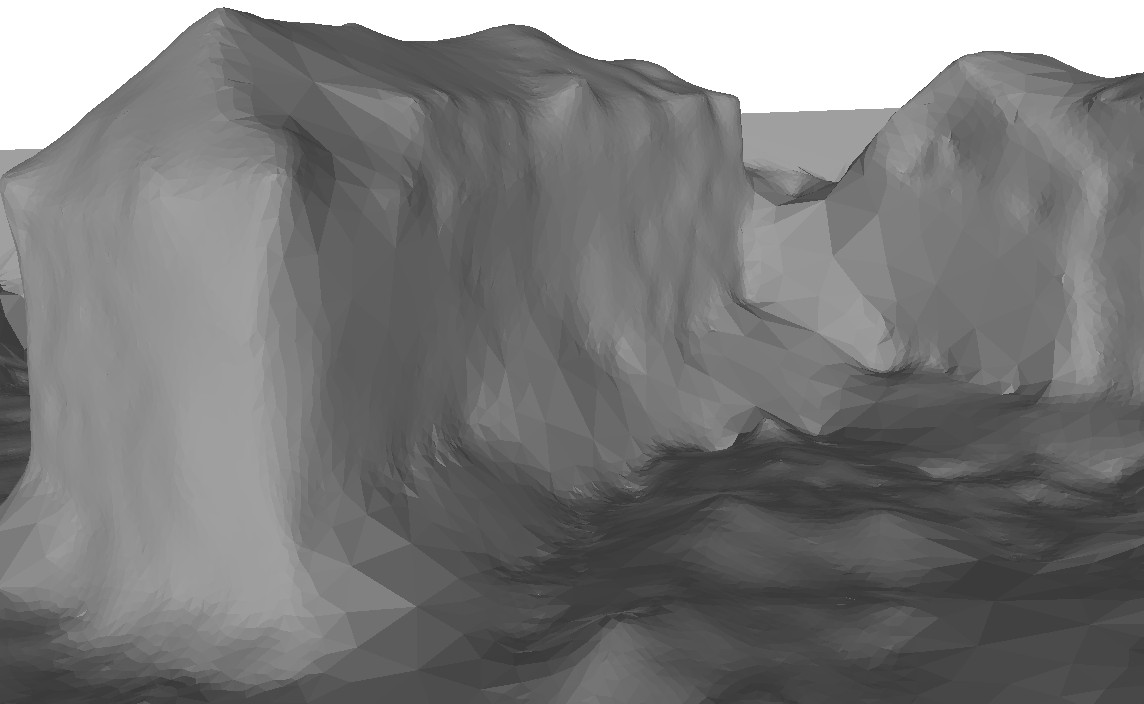}
\includegraphics[width=28mm, keepaspectratio]{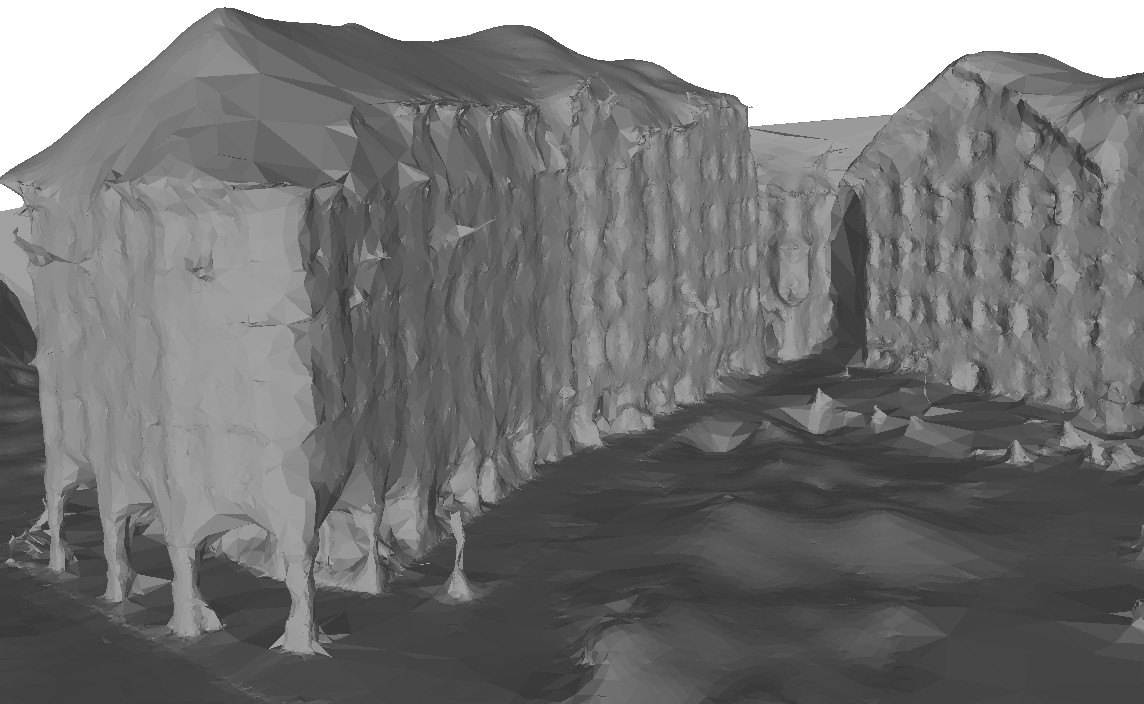}
\includegraphics[width=28mm, keepaspectratio]{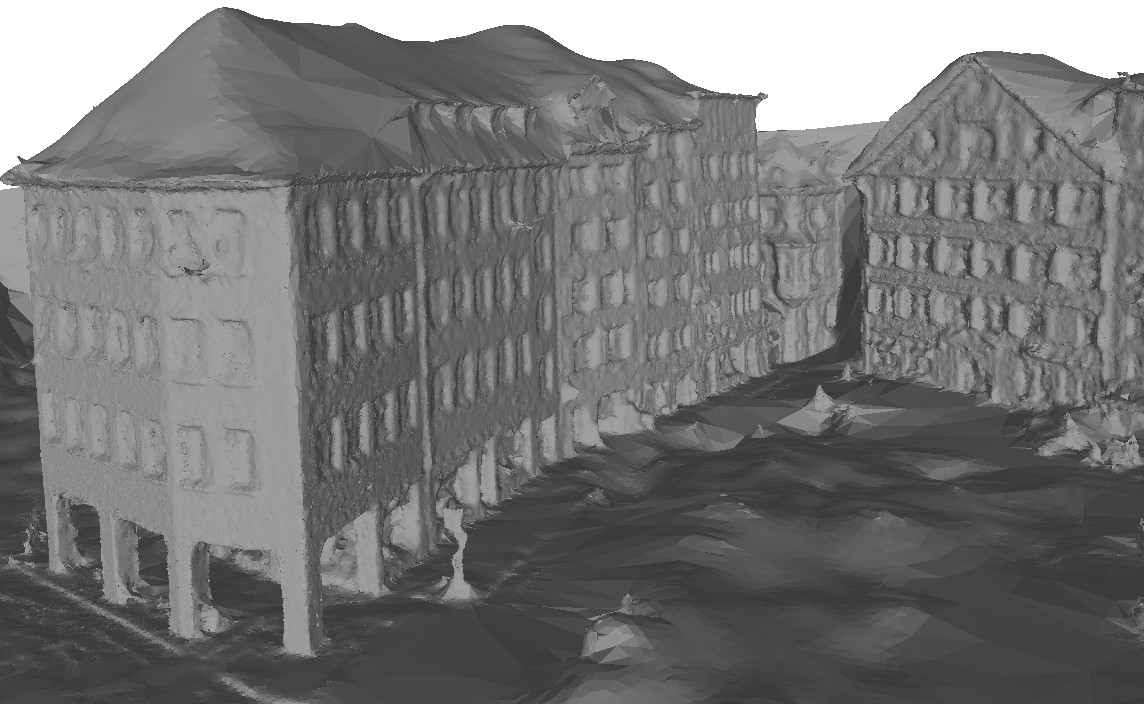}

\vspace{-3mm}
\caption{Our reconstruction of airborne-only (left) airborne + street-side SfM (middle) and airborne + street-side PMVS input (right).}
\label{fig:comparison}
\end{figure}

\begin{table*}[!ht]
 \centering
 \setlength{\tabcolsep}{0.8mm}
 \footnotesize
 \begin{tabular}{c|c|ccc|ccc|ccc|ccr|c|cccc|cccc}
 ~ & & \multicolumn{3}{c|}{Params} & \multicolumn{3}{c|}{Airborne data} & \multicolumn{3}{c|}{Street-side data} & \multicolumn{3}{c|}{3DT \& Ray shooting} & \multicolumn{1}{c|}{Mesh} & \multicolumn{4}{c|}{Mesh distance from ref (cm)} & \multicolumn{4}{c}{Timings}\\
 \hline
 & Experiment & vox & tr & $\lambda$ & \#pts & \#/m$^2$ & \#vis & \#pts & \#/m$^2$ & \#vis & \#verts & \#tets & \#rays & \#verts & mean$^a$ & mean$^s$ & 10cm$^s$ & 50cm$^s$ & tet & ray  & gco & total \\\hline\hline

 \parbox[t]{2mm}{\multirow{13}{*}{\rotatebox[origin=c]{90}{M\"unsterhof}}}
 &   PMVS (*) &   0  &    & 3 &    272k &    8.0 &   11.9 &   1.54M &   108.0 &    6.8 &   1.75M &    11.0M &    13.2M &    1.44M & (ref) & (ref) & (ref) & (ref) &     32s &     182s &      27s &     285s \\
 &   PMVS (a) &   0  &\chk& 3 &    272k &    7.9 &   11.9 &   1.54M &   106.8 &    6.8 &   1.75M &    11.0M &    13.2M &    1.43M & 2.9cm & 1.3cm & 0.7\% & 0.0\% &      32s &\bf{80s} &     28s & \bf{182s} \\
 &   PMVS (b) &   0  &    & 3 &    272k &    8.4 &\bf{1.0}&   1.54M &   108.9 &\bf{1.0}&   1.75M &    11.0M &\bf{1.75M}&    1.24M & 6.7cm & 1.5cm & 0.8\% & 0.0\% &      32s &\bf{28s} & \bf{34s}& \bf{137s} \\
 &   PMVS (c) &   0  &\chk& 3 &   272k  &    8.5 &\bf{1.0}&   1.54M &   108.5 &\bf{1.0}&   1.75M &    11.0M &\bf{1.75M}&    1.23M & 8.6cm & 1.6cm & 0.9\% & 0.0\% &      32s &\bf{15s} & \bf{34s}& \bf{125s} \\
 &   PMVS (d) & 0.10 &    & 3 &    272k &    8.4 &\bf{1.0}&\bf{780k}&\bf{56.3}&\bf{1.0}&\bf{996k}&\bf{6.27M}& \bf{996k}& \bf{756k}& 6.8cm & 1.7cm & 1.0\% & 0.0\% &  \bf{18s}&\bf{16s} & \bf{24s}& \bf{82s} \\
 &   PMVS (e) & 0.20 &    & 3 &    266k &    8.2 &\bf{1.0}&\bf{310k}&\bf{23.5}&\bf{1.0}&\bf{522k}&\bf{3.33M}& \bf{522k}& \bf{416k}& 7.1cm & 2.2cm & 2.5\% & 0.1\% & \bf{9.4s}&\bf{7.9s}& \bf{16s}& \bf{46s} \\
 &   PMVS (f) & 0.35 &    & 1 &\bf{213k}&\bf{6.3}&\bf{1.0}&\bf{127k}& \bf{6.3}&\bf{1.0}&\bf{293k}&\bf{1.89M}& \bf{293k}& \bf{262k}& 3.0cm & 3.1cm & 5.1\% & 0.3\% & \bf{5.2s}&\bf{4.2s}&\bf{6.0s}& \bf{23s} \\\cline{2-23}
    
 &    SfM (*) & 0 &      & 1 & 272k & 7.8 &    11.9  &  233k &  18.6 &     4.6  & 452k & 2.84M &   3.91M &    410k & 3.6cm & 7.7cm & 19.8\% & 1.4\% & 8.0s &     45s  & 7.6s &      70s\\
 &    SfM (a) & 0 & \chk & 1 & 272k & 7.6 &    11.9  &  233k &  18.7 &     4.6  & 452k & 2.84M &   3.91M &    412k & 3.6cm & 7.1cm & 18.1\% & 1.1\% & 9.8s & \bf{19s}  & 6.5s & \bf{47s}\\
 &    SfM (b) & 0 &      & 1 & 272k & 7.8 & \bf{1.0} &  233k &  18.7 & \bf{1.0} & 452k & 2.84M &\bf{452k}&\bf{382k}& 3.6cm & 7.5cm & 18.9\% & 1.3\% & 8.1s & \bf{6.4s} & 8.4s & \bf{33s}\\
 &    SfM (c) & 0 & \chk & 1 & 272k & 7.7 & \bf{1.0} &  233k &  19.0 & \bf{1.0} & 452k & 2.84M &\bf{452k}&\bf{383k}& 3.5cm & 7.5cm & 18.7\% & 1.3\% & 8.5s & \bf{3.2s} & 8.4s & \bf{29s}\\\cline{2-23}
 & aerial (*) & 0 &      & 1 & 272k & 6.2 &    11.9  &  \multicolumn{3}{c|}{not used}  &  272k &   1.77M &   3.24M &     269k  & 3.1cm & 69cm & 86.4\% & 49.1\% &      5.0s & 41s       & 3.3s & 52s \\
 & aerial (a) & 0 & \chk & 1 & 272k & 6.3 & \bf{1.0} &  \multicolumn{3}{c|}{not used}  &  272k &   1.77M &\bf{272k}& \bf{253k} & 3.1cm & 70cm & 86.5\% & 49.2\% & \bf{7.3s} & \bf{2.0s} & \bf{4.0s} & \bf{16s} \\\hline\hline

 \parbox[t]{2mm}{\multirow{10}{*}{\rotatebox[origin=c]{90}{Limmatquai}}}
 & PMVS (*) & 0.20 &    & 3 & 1.65M & 6.8 &   11.5 &   1.14M &   26.0 &   43.6 & 2.60M &    16.7M &     67.8M &    2.41M & (ref) & (ref) & (ref) & (ref) &     49s &   1059s &    53s &   1261s\\
 & PMVS (a) & 0.20 &\chk& 3 & 1.65M & 6.7 &   11.5 &   1.14M &   26.3 &   43.6 & 2.60M &    16.7M &     67.8M &    2.45M & 0.2cm & 0.3cm & 0.9\% & 0.1\% &     51s &\bf{314s}&    53s &\bf{517s}\\
 & PMVS (b) & 0.20 &    & 3 & 1.65M & 6.9 &\bf{1.0}&   1.14M &   26.5 &\bf{1.0}& 2.60M &    16.7M &     2.60M &\bf{2.11M}& 3.7cm & 2.6cm & 8.1\% & 0.3\% &     49s & \bf{50s}&\bf{79s}&\bf{277s}\\
 & PMVS (c) & 0.20 &\chk& 3 & 1.65M & 6.8 &\bf{1.0}&   1.14M &   26.9 &\bf{1.0}& 2.60M &    16.7M &     2.60M &\bf{2.13M}& 3.8cm & 3.0cm & 8.9\% & 0.4\% &     50s & \bf{19s}&\bf{78s}&\bf{248s}\\\cline{2-23}
 
 & SfM (*) &    0 &    & 1 & 1.68M & 6.5 &   11.5 &    259k &    9.1 &    4.6  & 1.79M &    11.5M &    19.6M &    1.72M & 0.3cm & 20cm & 44.4\% & 8.9\% &     35s &     262s &    35s &     380s\\
 & SfM (a) &    0 &\chk& 1 & 1.68M & 6.5 &   11.5 &    259k &    9.2 &    4.6  & 1.79M &    11.5M &    19.6M &    1.73M & 0.3cm & 19cm & 42.4\% & 7.7\% &     35s & \bf{95s} &    35s & \bf{213s}\\
 & SfM (b) &    0 &    & 1 & 1.68M & 6.6 &\bf{1.0}&    259k &    9.0 &\bf{1.0} & 1.79M &    11.5M &\bf{1.79M}&    1.62M & 0.6cm & 19cm & 44.0\% & 8.4\% &     34s & \bf{30s} &    39s & \bf{153s}\\
 & SfM (c) &    0 &\chk& 1 & 1.68M & 6.5 &\bf{1.0}&    259k &    9.5 &\bf{1.0} & 1.79M &    11.5M &\bf{1.79M}&    1.62M & 0.6cm & 21cm & 44.0\% & 8.8\% &     34s & \bf{12s} &    39s & \bf{134s}\\\cline{2-23}

 & aerial (*) &    0 &    & 1 & 1.68M & 6.0 &   11.5 &  \multicolumn{3}{c|}{not used}  & 1.68M & 10.9M &     19.3M  &     1.66M & 0.0cm & 81cm & 88\% & 47.3\% &  33s &    269s  &     33s  &     340s \\
 & aerial (a) &    0 &\chk& 1 & 1.68M & 6.0 &\bf{1.0}&  \multicolumn{3}{c|}{not used}  & 1.68M & 10.9M & \bf{1.68M} &     1.57M & 0.3cm & 81cm & 88\% & 47.4\% &  33s & \bf{12s} & \bf{26s} & \bf{103s} \\\hline

    \end{tabular}
	\caption{Summary of our results for two datasets. Changes w.r.t. the line (*) in each group are marked by bold text.}
\label{tab:results}
\vspace{-8mm}
\end{table*}

\begin{figure}[!ht]
\centering
\includegraphics[width=\linewidth, keepaspectratio]{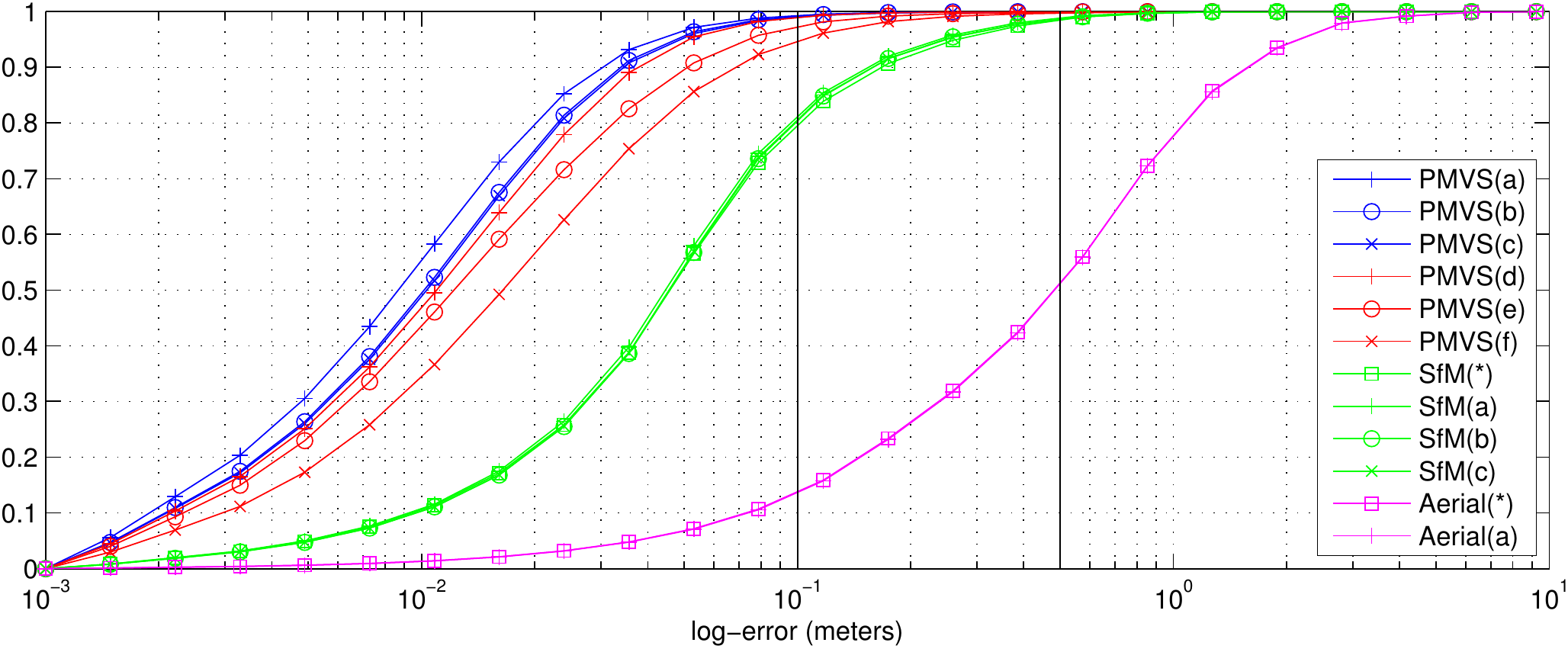}
\vspace{-7mm}
\caption{Cumulated distribution of mesh errors for the cases in Table~\ref{tab:results} over the street-side data regions of M\"unsterhof, the reference being PMVS(*). Vertical lines mark the thresholds 10cm and 50cm used for error reporting in the Table.}
\label{fig:cdf}
\vspace{-5mm}
\end{figure}

We applied our method to airborne-only data as a baseline and to airborne combined with street-side data from \emph{(i)} SfM and \emph{(ii)} PMVS and using different combinations of the reductions of Sect.~\ref{sec:simplify}. Fig.~\ref{fig:overview} shows different stages of our pipeline. We focus on the trade-off between runtime and output quality. Fig.~\ref{fig:comparison} compares models from different data types, while 
numerical results are summarized in Table~\ref{tab:results}. In all experiments, we fix our parameters as $\sigma_b=2$~m and $\lambda_b=1$ for point cloud blending, and $\sigma_{in}=0.1$~m, $\sigma_{out}=0.5$~m and $\gamma_{in}=\gamma_{out}=2$ for 3DT-fusion. The volumetric regularization parameter $\lambda$ is shown in Table~\ref{tab:results}. Denser input allows for better noise suppression without oversmoothing (higher $\lambda$). 
Input point decimation (Sect.~\ref{sec:simplify}) is parametrized by the voxel size \emph{vox} (0 for no decimation), ray truncation is marked by a \chk in column \emph{tr} and reduction of the number of rays is marked by value 1.0 in the columns reporting the number of visibility rays per point (\#vis) in Table~\ref{tab:results}. The input point density (\#/m$^2$) is estimated as the number of input points divided by output surface area. The table also reports the number of points (\#pts), tetrahedra (\#tets), and mesh vertices (\#verts) in each case. Note that the number of triangles in the 3DT (resp. output mesh) is nearly twice the vertex count. Since no detailed ground-truth 3D model exists for the fusion task to date, we evaluate the effect of the proposed simplifications w.r.t.~the best-quality 3D model obtained by our method (marked by \emph{(ref)} in Table~\ref{tab:results}). The distances from each vertex of the (dense) reference mesh to the output mesh is measured using AABB-trees \cite{CGAL}. As the aerial input is not altered in our experiments, and to report results independently from street-side coverage, we partition the reference mesh into street-side and aerial regions based on the origin of its vertices, and plot the error distributions for the street-side part in Fig.~\ref{fig:cdf}. Similarily, Table~\ref{tab:results} reports the mean error over the aerial (mean$^a$) and stree-side part (mean$^s$) separately, and the percentage of distances larger than 10 and 50~cm over the street-side part.

Runtimes are measured in Linux on a single i7-2600K 3.4GHz CPU core, 16 GB RAM. Table~\ref{tab:results} also lists runtimes for computing the 3DT and its adjacencies (tet), ray shooting (ray), energy computations and volumetric optimization (gco) and total time, which also includes point cloud $k$-NN and normal computation, point cloud blending and post-processing. 
For Limmatquai, we limited the input density to around 26 points/m$^2$ (via $vox=0.2$~m) having a memory peak of around 11.6 GB, the most memory-intensive steps being the volumetric decomposition and graph-cut optmization over it.

Our results in Table~\ref{tab:results} and Fig.~\ref{fig:cdf} consistently show that the quality of the model improves substantially by fusing street-side details into the airborne model, as expected. Improvements of using PMVS data instead of SfM are in the 1-10 cm range. Depending on the use-case, the quality provided by the street-side SfM may suffice. As an example, our method could be used for improving satellite or airborne Digital Surface Models (DSMs). It is also clear from the results that the proposed simplifications cause only little harm in quality (curves of the same color are close in Fig.~\ref{fig:cdf}) while a significant gain in runtime (and memory requirements) is achieved. We also found that ray truncation improves robustness to gross outliers. In particular, PMVS results in larger groups of outlier rays crossing the interior of certain buildings, while their effect remains local due to truncation. Although ray truncation makes the optimization local (unary terms are only non-zero in a crest around the data) \cite{LempitskyCVPR2007}, there are no serious consequences, once $\delta_{max}^{out}$ is set high enough so that street-side rays sweep out protrusions of the airborne input at the bottom of walls (see left of Fig.~\ref{fig:comparison}). As a failure case, point cloud blending tends not to disambiguate narrow structures (e.g.~towers) if these are duplicated distinctively due to misregistration in the input. 

\section{Conclusion}\label{sec:conclusion}
This paper proposes an efficient method for reconstructing a surface mesh from a detailed but incomplete street-side and a complete but low-detail airborne point cloud, which are pre-aligned. Although the next generation of 3D urban models should be both complete and contain street-side details, the problem received only little attention to date. Our method joins strengths of the two data types by fusing them via volumetric reasoning over a tetrahedralization, preceded by a point cloud blending to avoid gross line-of-sight conflicts due to inaccurate airborne measurements where more accurate street-side data is available. Our detailed experimentation with both SfM and dense MVS data over large urban scenes shows good surface quality and proves that additional simplifications can be used with little harm in output quality but substantial reduction in runtime. Applying the method at city scale via mobile mapping data is part of our future work. Ways to auto-adapt the parameters to input noise, ray density or tetrahedra volumes would also be interesting to study in the future.

\section*{Acknowledgment}
This work was supported by the European Research Council project VarCity~(\#273940). We thank Dr. Michal Havlena, Wilfried Hartmann and Prof.~Konrad Schindler for the Z\"urich aerial dataset, Michal Havlena for his comments on the paper.

\bibliographystyle{plain}
\bibliography{paper_fusion} 

\end{document}